\documentclass[10pt]{iopart}
\expandafter\let\csname equation*\endcsname\relax
\expandafter\let\csname endequation*\endcsname\relax
\usepackage{amsmath,amssymb,amsfonts}
\usepackage{graphicx}
\usepackage{textcomp}
\usepackage{wrapfig}
\usepackage{algorithmic}
\usepackage{algorithm}
\usepackage{array}
\usepackage{subcaption}
\usepackage{textcomp}
\usepackage{stfloats}
\usepackage{url}
\usepackage{verbatim}
\usepackage{graphicx}
\usepackage{cite}
\usepackage{enumitem}
\usepackage{subcaption}
\usepackage{multirow}
\usepackage{makecell}
\usepackage{booktabs}
\usepackage{tabularx}
\usepackage{cite}
\usepackage{caption}
\usepackage{lipsum}
\usepackage{xcolor}
\usepackage{hyperref}
\hypersetup{hypertex=true,colorlinks=true,linkcolor=blue,anchorcolor=blue,citecolor=blue}

\begin{document}

\title{2DLIW-SLAM:2D LiDAR-Inertial-Wheel Odometry with Real-Time Loop Closure}

\author{Bin Zhang$^{1,\ast}$ 
        Zexin Peng$^{1,\ast}$ 
        Bi Zeng$^{1,\dag}$ 
        and Junjie Lu$^{1}$ 
    \footnote{$^{\ast}$ Both authors contributed equally to the paper.
    }
    \footnote{$^{\dag}$ Corresponding author.}
}
\address{$^{1}$Guangdong University of Technology, Guangzhou, Guangdong, People’s Republic of China}

\ead{zomb2000@163.com}
\ead{zb9215@gdut.edu.cn}

\vspace{10pt}
\begin{indented}
\item[]April 2024
\end{indented}


\begin{abstract}
Due to budgetary constraints, indoor navigation typically employs 2D LiDAR rather than 3D LiDAR. However, the utilization of 2D LiDAR in Simultaneous Localization And Mapping (SLAM) frequently encounters challenges related to motion degeneracy, particularly in geometrically similar environments. To address this problem, this paper proposes a robust, accurate, and multi-sensor-fused 2D LiDAR SLAM system specifically designed for indoor mobile robots. To commence, the original LiDAR data undergoes meticulous processing through point and line extraction. Leveraging the distinctive characteristics of indoor environments, line-line constraints are established to complement other sensor data effectively, thereby augmenting the overall robustness and precision of the system. Concurrently, a tightly-coupled front-end is created, integrating data from the 2D LiDAR, IMU, and wheel odometry, thus enabling real-time state estimation. Building upon this solid foundation, a novel global feature point matching-based loop closure detection algorithm is proposed. This algorithm proves highly effective in mitigating front-end accumulated errors and ultimately constructs a globally consistent map. The experimental results indicate that our system fully meets real-time requirements. When compared to Cartographer, our system not only exhibits lower trajectory errors but also demonstrates stronger robustness, particularly in degeneracy problem.
\end{abstract}

\vspace{2pc}

\submitto{}

\maketitle

\ioptwocol

\section{Introduction}
SLAM plays a crucial role in facilitating autonomous localization and map construction for robots in unknown environments, thereby providing substantial support for robot autonomy and navigation\cite{kolhatkar_review_2021}. As SLAM continues to advance, its real-life applications have garnered increasing interest for enhancing the quality of life and fostering a smarter society. Particularly in indoor environments, where GPS signals often prove unreliable for robot localization, SLAM technology proves highly suitable for efficient positioning.

At present, mainstream indoor mobile robots predominantly rely on 2D LiDAR for localization and mapping, such as floor-cleaning robots and restaurant service robots. However, a limitation of these robots lies in their loose coupling with other sensors. For instance, Cartographer\cite{hess_real-time_2016} fails to fully leverage the potential performance of each sensor, potentially resulting in motion degradation when the environmental geometric structure is relatively simple. Additionally, due to sensor constraints, 2D LiDAR SLAM typically assumes strict horizontal ground surfaces during modeling, effectively restricting the robot's motion to three degrees of freedom (3DoF). Although this assumption holds true in many indoor scenarios, it may not be applicable in complex environments.

In scenarios where six degrees of freedom (6DoF) robotic motion is applicable, sensor combinations involving 3D LiDAR and an Inertial Measurement Unit (IMU), such as FAST-LIO\cite{xu_fast-lio_2021}, LIO-SAM\cite{shan_lio-sam_2020}, have shown promise. While 3D LiDAR can provide 3D point cloud data, leading to higher accuracy in the optimization process, this comes at the cost of consuming more computational resources and requiring more expensive hardware compared to 2D LiDAR, as also mentioned in BLVI-SLAM\cite{liuFusionBinocularVision2022}. As a result, many indoor robots are equipped solely with 2D LiDAR, which provides only 2D scanning data. Our aim is to extract geometric structures, such as corner points and straight lines, from 2D scan data for tracking and loop detection, similar to 3D LiDAR front-end processing. Additionally, the motion of most indoor mobile robots typically exhibits approximately 3DoF, with only minor variations occurring in a few axes, such as roll and pitch. This suggests the possibility of estimating 6DoF using a 2D LiDAR, making it a viable primary sensor choice.

In recent times, propelled by advancements in Visual SLAM\cite{liuVisualSLAMMethod2023} and 3D LiDAR SLAM\cite{heTightlyCoupledLaserinertial2023}, multi-sensor fusion for motion and pose estimation has found extensive utility. Given that indoor robots generally move at slow speeds with frequent stops and starts, the performance of IMU under such conditions may not be ideal, whereas wheel odometry demonstrates superior performance, especially when the robot is stationary and there is no error accumulation. Moreover, as most indoor robots are wheeled robots, their motion state can be inferred to some extent through encoder information. However, the fusion of 2D LiDAR and wheel odometry remains relatively underexplored in this domain.

Considering these factors, relying solely on a combination of 2D LiDAR and IMU might not offer sufficient robustness, particularly in scenarios featuring significant dynamic interference and degraded environments. Such conditions can lead to erroneous motion state estimates, potentially resulting in system crashes. Thus, a more reasonable approach involves augmenting the sensor combination by integrating the robot's built-in wheel odometry, representing a cost-effective solution. Hence, the primary objective of this paper is to design a fast, stable, and cost-effective 2D LiDAR-inertial-wheel odometry. Furthermore, our proposed system incorporates real-time loop detection and pose graph optimization to create a comprehensive SLAM system, termed as 2DLIW-SLAM.

The primary contributions of this paper are as follows:
\begin{itemize}
    \item Development of a novel point-line feature extraction method tailored for 2D LiDAR data, along with the construction of a corresponding relative pose transformation observation model. Upon that, we present a multi-sensor tightly-coupled odometry that seamlessly fuses data from 2D LiDAR, IMU, and wheel odometry.
    \item Introduction of a global feature-based loop detection algorithm that achieves high-precision matching while meeting real-time requirements. 
    \item We incorporate additional constraints, such as ground constraints, and integrate pivotal components such as initialization, graph optimization, and probability grid map, among others, thus assembling a complete SLAM system.
\end{itemize}
\begin{figure*}
    \centering
    \includegraphics[width=1\linewidth]{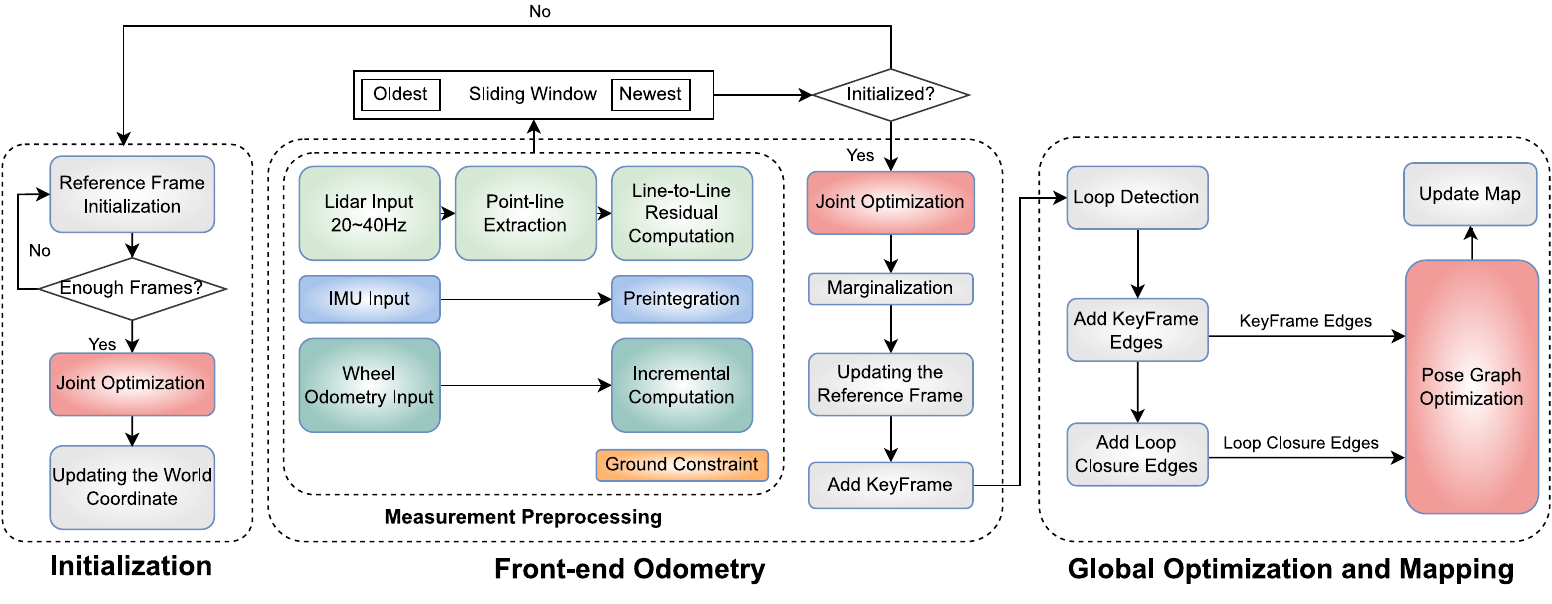}
    \caption{The system structure of 2DLIW-SLAM. The Initialization and Front-end Odometry will be elaborated in detail in Section IV. The Global Optimization and Mapping will be discussed comprehensively in Section V.}
    \label{fig:overview}
\end{figure*}
\section{Related Work}
\textbf{LiDAR Odometry and Mapping.} Before the optimization algorithm, SLAM heavily relied on filtering algorithms\cite{steux_tinyslam_2010,miller_rao-blackwellized_2007,grisetti_improved_2007}.  Gmapping\cite{grisetti_improved_2007} further enhanced filtering performance by reducing the number of particles and minimizing resampling frequency. Subsequently, graph optimization-based techniques began to gain prominence\cite{konolige_efficient_2010,durrant-whyte_linear_2012,hess_real-time_2016}. Notably, Cartographer\cite{hess_real-time_2016} incorporates the concept of submaps and leverages 2D LiDAR for loop closure detection, thus effectively addressing long-term drift issues. 

Subsequently, the primary research focus shifted towards 3D LiDAR SLAM. Zhang \textit{et al.}\cite{zhang_loam_2014} introduced LOAM, a method leveraging geometric features in the environment to establish LiDAR odometry based on point-line and point-plane constraints, significantly enhancing the accuracy of LiDAR odometry. Additionally, they introduced loop closure detection functionality, culminating in a comprehensive SLAM framework called LeGO-LOAM\cite{shan_lego-loam_2018}. 
Following this, by integrating IMU data and applying ISAM2\cite{kaess_isam2_2011} theory, they proposed LIO-SAM\cite{shan_lio-sam_2020}, resulting in further system performance enhancement.

In addition to the aforementioned approaches, Wei \textit{et al.}\cite{xu_fast-lio_2021,xuFASTLIO2FastDirect2022} proposed the FAST-LIO and FAST-LIO2 series of methods, which effectively fuse 3D LiDAR and IMU data using the Iterated Error Kalman Filter (iEKF) and have achieved remarkable results.

Recent developments in the SLAM field have gravitated toward multi-sensor fusion, leading to the emergence of SLAM frameworks that integrate LiDAR, IMU, and visual data. Noteworthy examples include R2LIVE\cite{lin_r_2021} and R3LIVE\cite{lin_r3live_2022} proposed by Lin \textit{et al.}, and LVIO-SAM\cite{zhong_lvio-sam_2021} proposed by Zhong \textit{et al.}. Zhang \textit{et al.} \cite{zhangAccurateRealtimeSLAM2022} have improved traditional point cloud registration and loop-detection methods, proposing a new scheme for multi-sensor SLAM systems. GLIO \cite{liuGLIOTightlyCoupledGNSS2024} tightly couples GNSS, LiDAR, and IMU integration for continuous and drift-free state estimation of intelligent vehicles in urban areas. Song \textit{et al.} \cite{songReliableEstimationAutomotive2023} utilized a dual neural network and a square-root cubature Kalman filter to mitigate sensor noise and varied maneuvers. These approaches further enhance localization accuracy and robustness by leveraging multiple sensor modalities.

\textbf{Wheel Odometry in SLAM.} \cite{zhang_LiDAR-imu_2019} represents the first attempt to fuse LiDAR, IMU, and wheel odometry in a loosely coupled manner. Initially, specific states are computed using data from the LiDAR, IMU, and wheel odometry. Subsequently, these states are integrated into an Extended Kalman Filter (EKF), resulting in the final state estimation. EKF-LOAM\cite{junior_ekf-loam_2022} also adopts the EKF framework and incorporates a simple and lightweight adaptive covariance matrix based on the detected geometric feature count. The OpenLORIS-Scene\cite{shi_are_2020} dataset provides wheel odometry data for experimentation. However, the incorporation of wheel odometry in SLAM is not yet widespread, and open-source implementations are relatively scarce. \cite{shi_are_2020} also emphasizes that SLAM algorithms should not overlook the advantages and benefits offered by wheel odometry.

\textbf{Loop Detection.} Loop detection is a critical process in SLAM. Cartorgrapher\cite{hess_real-time_2016} employs a search window-based technique to match the current LiDAR frame with the map for loop closure detection. Kim \textit{et al.}\cite{kim_scan_2018} encode a LiDAR frame into rotation-invariant descriptors and accelerates the matching process using the KD-Tree method. Building upon this algorithm, a descriptor with both rotation and translation invariance, called scancontext++\cite{kim_scan_2022}. Xiang \textit{et al.}\cite{xiang_fastlcd_2021} combined encoding and machine learning to propose FastLCD for handling point cloud data and achieving reliable and accurate loop closure detection. OverlapTransformer\cite{ma_overlaptransformer_2022} uses a lightweight neural network to expose the range image representation of point clouds, enabling fast retrieval. BoW3D\cite{cui_bow3d_2023} presented a bag-of-words model applicable to 3D LiDAR data, which not only effectively identifies previously visited locations but also real-time corrects the complete pose.

\section{Our System Overview}
We consistently employ rotation vector $\theta$ to represent pose and $p$ to denote the translation vector. The optimized state variables include the robot's position $p_k \in \mathbb{R}^3$ and orientation angles  $\theta_k \in \mathrm{so}(3)$ in the world coordinate system, velocity $v_k \in \mathbb{R}^3$, as well as acceleration bias $b_{\alpha_k} \in \mathbb{R}^3$ and gyroscope bias $b_{\omega_k} \in \mathbb{R}^3$:
\begin{equation}
    \begin{aligned}
        X=[p_k,\theta_k,v_k,b_{\alpha_k},b_{\omega_k}]^T
    \end{aligned}
\end{equation}

A general SLAM structure is adopted, comprising two main components: front-end odometry and back-end optimization, as depicted in figure~\ref{fig:overview}. 

The front-end odometry tighly couples three primary sensors: a 2D LiDAR, an IMU, and wheel odometry. The 2D LiDAR is used to extract line features and optimizes the current frame's pose by matching these features with those from the reference frame. The IMU adopts an IMU pre-integration observation model. On the other hand, the wheel odometry captures changes in motion over a period, with more detailed information provided later in this paper. The implementation of ground constraints ensures that the robot remains well-constrained even under slight perturbations, effectively reducing the robot's 6DoF to 3DoF. Additionally, the odometry periodically transmits keyframes to the back-end.

The back-end primarily consists of loop detection and pose graph optimization. Initially, it receives keyframes from the front-end and establishes relative pose constraints between them. Periodic loop closure detection is performed to establish loop closure constraints in the pose graph, which is then subjected to global optimization. Finally, keyframes will be employed to periodically create a two-dimensional probability grid map.

\section{Front-end odometry}
\subsection{Extracting points and lines}
\begin{figure}
    \centering
    \includegraphics[width=0.6\linewidth]{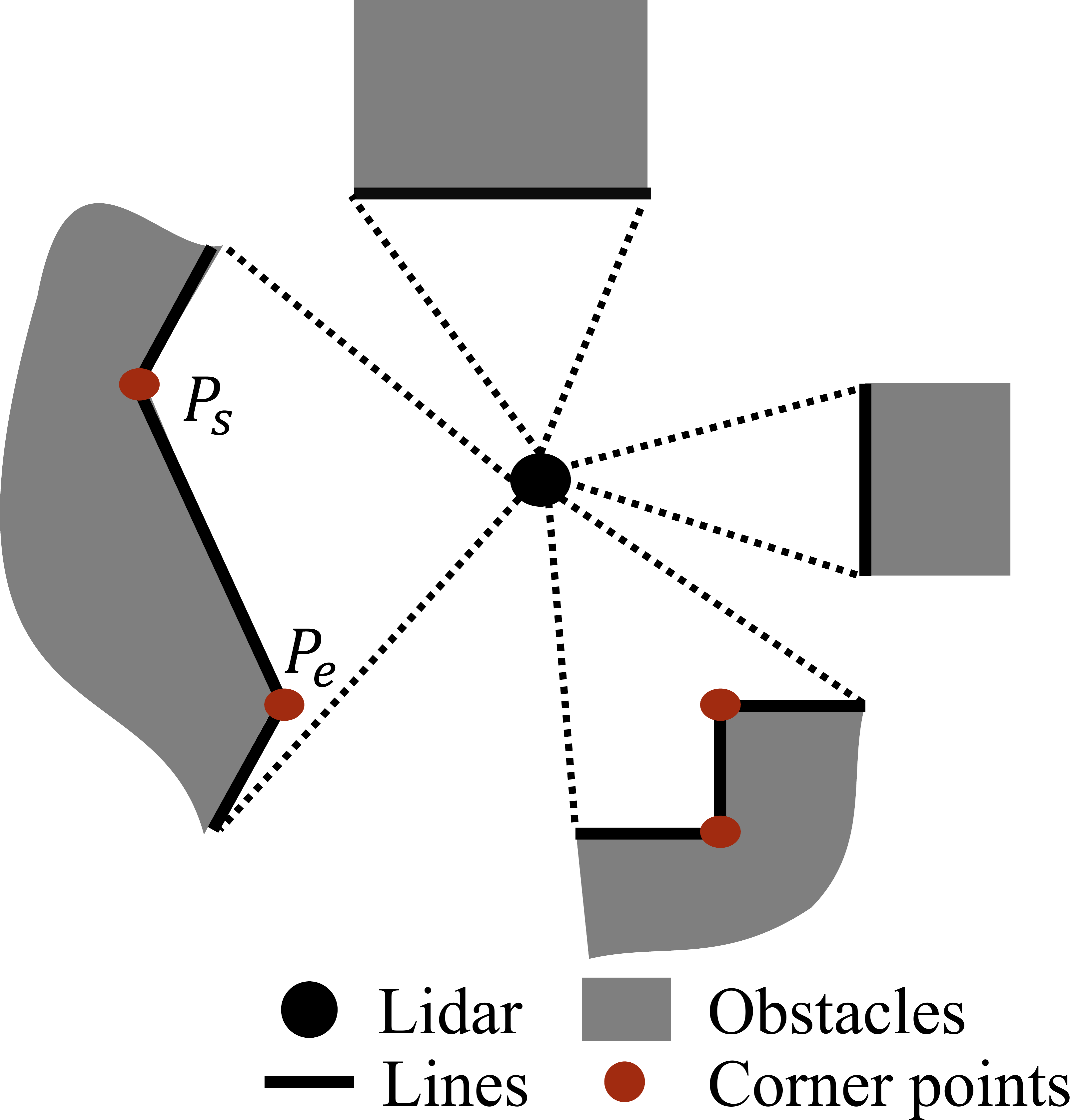}
    \caption{Illustration of line feature extraction. All points are processed using  $A^TP=0$ to obtain the linear equation, and the line segment range is defined by $(P_{s},P_{e})$.}
    \label{fig:extract_features}
\end{figure}

The point-line process employed in this paper consists of two main steps: line extraction and corner extraction based on the identified lines, as depicted in figure~\ref{fig:extract_features}. The extracted lines are utilized for front-end state estimation, while the corners play a crucial role in back-end mapping for loop closure detection. The specific point-line extraction process is detailed as follows:

Initially, the continuity is determined based on the distance between neighboring points, so that a scan of points is divided into multiple continuous point sets $\{S_1,...,S_n\}$.

Each point set $S_i$ is processed independently. For every point $P_i$, the angle $\theta$ formed by the two vectors connecting $P_i$ to its previous point $P_{i-1}$ and next point $P_{i+1}$ are computed. 
\begin{equation}\label{eq:cal_angle}
    \begin{aligned}
        l_i &= P_i-P_{i-1},l_j=P_{i+1}-P_i \\
        \theta_i &= \arccos{\frac{l_il_j}{\|l_jl_j\|}}.
    \end{aligned}
\end{equation}
Due to the small jitter between neighboring points in a scan of points, simply using neighboring points will be more affected by noise and result in larger fluctuations. Therefore, when selecting neighboring points, points that are too close together are skipped. For each $\theta$, we use Non-Maximum Suppression (NMS) to determine the largest angle in the retained neighborhood.

Starting from the initial point $P_s$ of $S_i$, we use equation~\eqref{eq:cal_angle} to calculate $\theta$ formed by $P_s$, the current potential endpoint $P^c_e$ and the next potential endpoint $P^n_e$. If $\theta$ exceeds a set threshold, they are considered to be in the same line. This process continues until one $\theta$ below the threshold is encountered or the endpoint is reached. 

All points between $P_s$ and the final endpoint $P_e$ are treated as belonging to the same line, and a least squares fitting is applied to all points within this range.
\begin{equation}
    \begin{aligned}
    \left[\begin{array}{ccc}x_1&y_1&1 \\ \vdots&\vdots&\vdots \\ x_n&y_n&1\end{array}\right]
    \left[\begin{array}{c}a \\ b \\ c\end{array}\right]=
    \left[\begin{array}{c}P_1 \\ \vdots \\ P_n\end{array}\right]A=0       
    \end{aligned}
\end{equation}
where the line expression $A^TP=0$ is derived, enabling the simultaneous calculation of the projection coordinates $P_{s}$ and $P_{e}$ for the two endpoints on this line. Throughout this paper, line segments are represented using the tuple $(P_{s},P_{e})$. 

Once a line segment is identified, the current point is designated as the new $P_s$, and the aforementioned process is reiterated until all potential endpoints are processed. To assess the continuity of a detected line segment with adjacent lines, the angle between the two lines is computed. If this angle surpasses a specific threshold, it is regarded as a stable corner point formed by the intersection of the two lines. Using the aforementioned algorithm, stable lines and corners can be effectively extracted. To facilitate rapid subsequent matching, a map is constructed for each scan frame. 

\subsection{Line-to-Line Alignment Constraints}
\begin{figure}
    \centering
    \includegraphics[width=0.7\linewidth]{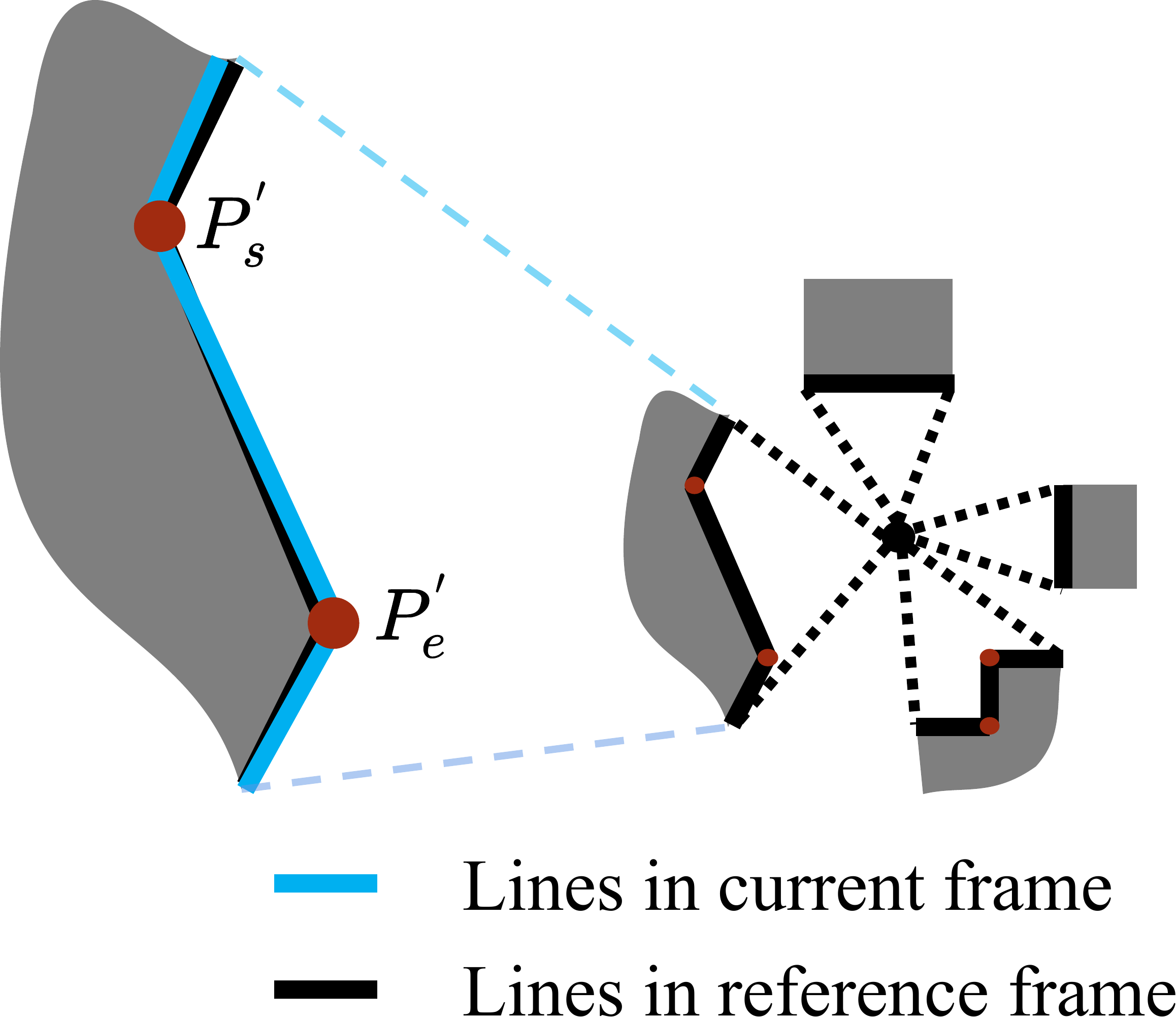}
    \caption{Illustration of line feature matching optimization between the current frame and the reference frame. The distances between $(P^{'}_{s},P^{'}_{e})$ and line in the reference frame are utilized as observational constraints to optimize the pose of the current frame.}
    \label{fig:match_line}
\end{figure}

By leveraging wheel odometry information and past state estimation, we can derive the pose change during this period, represented as $T^{O_r}_{O_c}$. Furthermore, based on the relative pose $T^O_L$ between the LiDAR and the wheel odometry, we calculate the pose change of the LiDAR between the current frame and the reference frame, denoted as $T^{L_r}_{L_c}$. We adopt a specific strategy to select reference frames from historical frames. After that, the extracted lines are compared with line data from the reference frame, which is depicted in figure~\ref{fig:match_line}.

Utilizing $T^{L_r}_{L_c}$, we effectively map the lines $l_c=(P_s, P_e)$ in the current scan to the reference frame $L_r$. In other words, we establish their corresponding positions and orientations in the reference frame.
\begin{equation}
    l^{'}_c=(P^{'}_s,P^{'}_e)=(T^{L_r}_{L_c}P_s,T^{L_r}_{L_c}P_e).
\end{equation}
Upon identifying these lines, they undergo a threshold check: if the angle between any two lines exceeds a certain threshold, they are discarded from consideration. In cases where multiple lines meet the criteria, the line with the smallest angle is selected as the matched line $l_r$, thereby forming a line matching pair $l_{m_i}=(l_r,l_c)$. By systematically finding all possible line-line matching pairs between the current frame $L_c$ and the reference frame $L_r$, a set of matching lines is obtained, denoted as $m=\{ l_{m_i}=(l_r,l_c) \}$.

Given that individual lines generally do not strictly overlap, we compute the distance $d_1$ from the initial point $P^{'}_s$ to the target line $l_r$ and $d_2$ from the terminal point $P^{'}_e$ to $l_r$\cite{zhang_loam_2014}. Specifically, this constraint can be expressed as follows:
\begin{align}
    d_1 &= \left\Vert (P^{l_r}_s-P_e) - \left( \frac{P^{l_r}_e-P^{l_r}_s}{\| P^{l_r}_e-P^{l_r}_s \|} \right)^T \right. \nonumber \\
    &\qquad \left. \cdot (P^{l_r}_s-P_e) \frac{P^{l_r}_e-P^{l_r}_s}{\| P^{l_r}_e-P^{l_r}_s \|} \right\Vert.
\end{align}
Likewise, $d_2$ is computed. With both $d_1$ and $d_2$ determined, the observation model can now be formulated as follows:
\begin{equation}\label{eq:line_obs}
    h_L(p_r,\theta_r,p_c,\theta_c)=d_1+d_2+n_l=0
\end{equation}
where it is assumed that $n_l$ follows Gaussian noise with a distribution of $n_l \sim \mathcal{N}(\mu, \sigma^2_l)$. This observation model will be integrated into the joint optimization process in the following steps.

\subsection{IMU Preintegration}
Since the state at time $t_{B_k}$ changes continuously in nonlinear optimization processes, IMU state propagation must be performed frequently, leading to a significant performance loss. To overcome this issue, we adopt the IMU preintegration\cite{qin_vins-mono_2018}:
\begin{equation}\label{eq:6}
\begin{aligned}
    \alpha^{B_k}_{B_{k+1}} &=
        \iint_{t_{B_k}}^{t_{B_{k+1}}} R^{B_k}_t(\hat{a_t}-b_{a_t}-n_a)  \, dt^2 \\
    \beta^{B_k}_{B_{k+1}} &=
        \int_{t_{B_k}}^{t_{B_{k+1}}} R^{B_k}_t(\hat{a_t}-b_{a_t}-n_a)  \, dt \\
    \gamma^{B_k}_{B_{k+1}} &=
        I + \int_{t_{B_k}}^{t_{B_{k+1}}} \gamma^{B_k}_{t}\lfloor \hat{\omega}-b_{\omega_t} -n_{\omega}\rfloor_{\times} \, dt.
\end{aligned}
\end{equation}
As observed, this integrated part depends solely on the IMU bias values. Finally, the IMU observation model can be derived as follows:
\begin{equation}\label{eq:IMU}
\begin{aligned}
    &\begin{bmatrix}
        R^{B_k}_W (p^W_{B_{k+1}} - p^W_{B_k} + v^W_{B_K}\Delta t^2) - \alpha^{B_k}_{B_{k+1}} \\ 
        R^{B_k}_W (v^W_{B_{k+1}} - v^W_{B_k} + g^W\Delta t) - \beta^{B_k}_{B_{k+1}} \\ 
        \exp{(\theta^{w}_{b_{k+1}})}^T \exp{(\theta^{w}_{b_{k+1}})} \gamma^{B_k}_{B_{k+1}} \\ 
        b_{a_{B_{k+1}}} - b_{a_{B_{k}}} \\ 
        b_{\omega_{B_{k+1}}} - b_{\omega_{B_{k}}}
    \end{bmatrix} + n_B \\
    &= h_B(p^W_{B_k},\theta^W_{B_k},v^W_{B_k},p^W_{B_{k+1}},\theta^W_{B_{k+1}},v^W_{B_{k+1}}) + n_B.
\end{aligned} 
\end{equation}
Equation~\eqref{eq:IMU} establishes the relationship between the position, orientation, and velocity of the robot in the current frame $B_k$ and the neighboring frame $B_{k+1}$, along with the corresponding noise term $n_B$. The noise term $n_B$ follows Gaussian noise, \textit{i.e.}, $n_B \sim \mathcal{N}(\mu, \Sigma_B)$.

\subsection{Observation of Wheel Odometry}
We adopt the assumption that the data acquired directly from the wheel odometry represents the change in the chassis pose over a period, denoted as $\hat{T}^{O_i}_{O_{i+1}}$. This can be calculated using the values $p^W_{B_k},\theta^W_{B_k},p^W_{B_{k+1}},\theta^W_{B_{k+1}}$.

Furthermore, the wheel odometry data can be transformed into translational magnitude $d$, directional change $\theta_d$, and pose variation $\theta$ as follows:
\begin{equation}
    \begin{aligned}
        \hat{d}^i_{i+1}&=\|\hat{p}^i_{i+1}\| \\
        (\hat{\theta}_d)^i_{i+1}&=atan\left(\frac{(\hat{p}^i_{i+1})_y}{(\hat{p}^i_{i+1})_x} \right) \\
        \hat{\theta}^i_{i+1}&=\|\log(\hat{R}^i_{i+1})\|
    \end{aligned}
\end{equation}
where $\hat{d}^i_{i+1}$, $(\hat{\theta}_d)^i_{i+1}$, $\hat{\theta}^i_{i+1}$, and 
$\hat{R}^i_{i+1}$ represent the magnitude of the wheel odometry displacement increment, the angle of displacement increment, the angle of pose change, and the rotation matrix, respectively.

By utilizing these equations, the observation model can be derived as follows:
\begin{equation}\label{eq:11}
    \begin{aligned}
        &\begin{bmatrix}
        \hat{d}^{O_k}_{O_{k+1}}-d^{O_k}_{O_{k+1}} \\ 
        \hat{\theta}^{O_k}_{d_{O_{k+1}}}-\theta^{O_k}_{d_{O_{k+1}}} \\ 
        \hat{\theta}^{O_k}_{O_{k+1}}-\theta^{O_k}_{O_{k+1}}
        \end{bmatrix} + n_O \\
        = &h_O(p^W_{B_k},\theta^W_{B_k},p^W_{B_{k+1}},\theta^W_{B_{k+1}}) + n_O.
    \end{aligned}   
\end{equation}
where $n_O$ follows Gaussian noise, \textit{i.e.}, $n_O \sim \mathcal{N}(\mu, \Sigma_O)$, with its covariance matrix computed simultaneously during the wheel odometry increment extraction process.
\subsection{Ground Constraint}
The wheel odometry mainly provides motion information in the $x$-$y$ plane. To achieve a reliable estimation of the 6DoF state, it becomes essential to incorporate ground constraints to stabilize the system. 

Given the IMU's pose in the world coordinate system, denoted as $p^W_{B_k}, \theta^W_{B_k}$, we compute the pose of the wheel odometry in the world coordinate system, represented as $p^W_{O_k}, \theta^W_{O_k}$, using the following equation:
\begin{equation}\label{eq:12}
    \begin{aligned}
        (p^W_{O_k},\theta^W_{O_k})=\log{\left( \exp{ \left( p^W_{B_k},\theta^W_{B_k} \right) T^B_O }\right)}.
    \end{aligned}
\end{equation}
using equation~\eqref{eq:12}, we can derive the direction vector of the $z$-axis in the current wheel odometry frame and its value in the world coordinate system, expressed as:
\begin{equation}
    \begin{aligned}
        axis_z &= \left( \exp{(\theta^W_{O_k})} \right)_z \\
        z &= (p^W_{O_k})_z
    \end{aligned}   
\end{equation}
where $(\cdot)$ denotes the third element of the vector. $axis_z$ and $z$ represent the orientation and value of the$z$-axis in the current wheel odometry frame, respectively.

We establish an observation model as follows:
\begin{equation}\label{eq:14}
    \begin{aligned}
       &\begin{bmatrix} z \\ 
            \| \arcsin(axis_z \times \begin{bmatrix} 0 & 0 & 1 \end{bmatrix}^T) \|  
        \end{bmatrix} + n_g \\
        &= h_g(p^W_{B_k},\theta^W_{B_k}) + n_g
    \end{aligned}
\end{equation}
where $n_g$ follows Gaussian distribution, i.e., $n_g \sim \mathcal{N}(0,\Sigma_g)$, with the actual covariance depending on the level of ground jitter. By employing the ground constraint, the robot's motion is constrained to 3DoF, which aligns better with real-world observations.

\subsection{Joint Optimization}
By applying Maximum Likelihood Estimation (MLE) to the observations presented in equation~\eqref{eq:line_obs}, equation~\eqref{eq:IMU}, equation~\eqref{eq:11} and equation~\eqref{eq:14}, we can effectively solve the following nonlinear problem:
\begin{equation}\label{eq:15}
    \begin{aligned}
        X^{*}&=\mathop{ \arg\min\limits_{X}{ \sum_{i=k-w}^k\sum_{j=1}^{m_i} \| h_{L_j}(p_r,\theta_r,p_i,\theta_i) \|^2_{\sigma^2_l} } }  \\
        &+ \|h_B(p^W_{B_{k-1}},\theta^W_{B_{k-1}},v^W_{B_{k-1}},p^W_{B_{k}},\theta^W_{B_{k}},v^W_{B_{k}})\|_{\Sigma_B}\\
        &+ \|h_O(p^W_{B_{k-1}},\theta^W_{B_{k-1}},p^W_{B_{k}},\theta^W_{B_{k}})\|_{\Sigma_O} \\
        &+ \sum^k_{i=o}\|h_g(p^W_{B_k},\theta^W_{B_k}) \|_{\Sigma_g}
    \end{aligned}
\end{equation}
where the subscript $k$ designates the current frame, while $k-1$ corresponds to the previous frame and $w$ denotes the sliding windows size.

In the front-end odometry tracking process, prior factors are introduced simultaneously to aid in the estimation.

\subsection{Initialization}
Before optimizing the position of the front-end odometry, we first perform an initialization of the system. This process involves setting the reference frame and using the line features and corner point features of the LiDAR frame in the initialization window to obtain a better initial position. The specific steps involved are as follows:
\begin{enumerate}
    \item Accumulate n consecutive non-stationary LiDAR frames, and when a certain quantity criterion k is met, the system initiates the initialization check.\label{item:step1}
    \item If each frame can establish a sufficient number of line-to-line correspondences with the first frame's LiDAR lines, the formal initialization process begins. Otherwise, all data is cleared, and the initialization preparation returns to Step~\ref{item:step1}.
    \item Perform joint optimization based on equation~\eqref{eq:15}.
\end{enumerate}

\subsection{Construction of Reference Frames}
\begin{enumerate}
    \item The system maintains two reference frames: the current reference frame and the backup reference frame. When the current reference frame is empty, the current frame becomes the reference frame.\label{item:1}
    \item Estimate the state of the current frame and map the line data of the current frame onto the current reference frame based on the estimation result. If the corresponding cell already contains line data, the cell remains unchanged; otherwise, the cell is updated.
    \item When the accumulated data in the reference frame reaches half of the threshold, the current frame data is synchronously updated to both the current reference frame and the backup reference frame.
    \item If the current reference frame reaches the threshold, it is deleted, and the backup reference frame takes its place as the current reference frame. A new empty backup reference frame is then created. Subsequently, repeat Steps \ref{item:1} to \ref{item:4}. \label{item:4}
\end{enumerate}

\section{Global Optimization and Mapping}
\subsection{Loop Detection}
\begin{figure}
    \centering
    \includegraphics[width=1.0\linewidth]{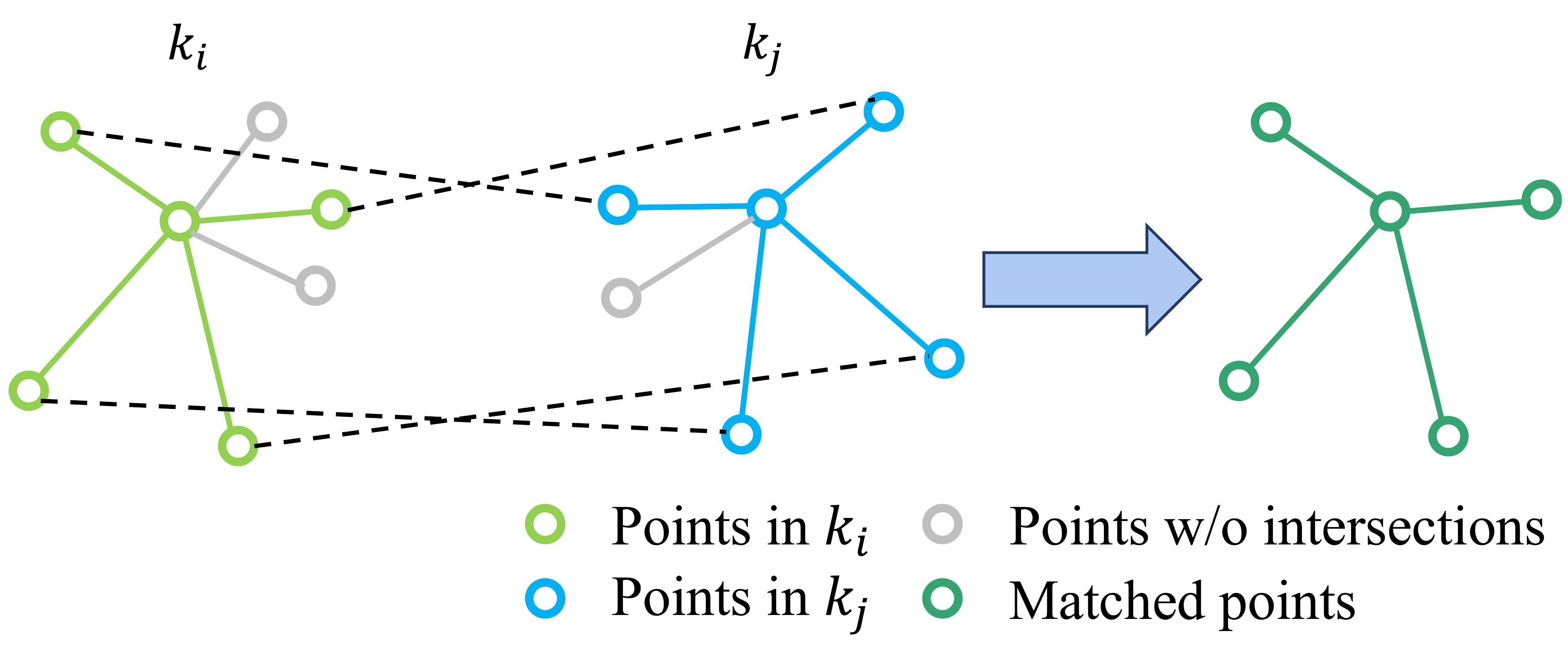}
    \caption{Illustration of loop closure matching. All points are corner feature points, and similarity matching is performed by comparing the distance and angle features between these corner points.}
    \label{fig:loop}
\end{figure}

2D LiDAR scan describes the geometric structure of the environment, e.g., straight lines on walls, corners of rooms. The current common feature descriptors such as ORB\cite{zeng_orb-slam2_2018} or SFIT\cite{loweDistinctiveImageFeatures2004} are difficult to characterize such 2D scans that lack texture features. Therefore, we have developed a method for matching global feature point descriptors that can characterize 2D scan data based on its geometric structure. We employ this method for loop detection, as depicted in~\ref{fig:loop}. In our front-end processing pipeline, we extract corner features from the space between two line segments. These corner features will be employed for loop closure detection. Consequently, corresponding descriptor sets are generated for each keyframe, ensuring adequate information to address the issue of insufficient data in individual frame scans. 

Based on the stability of corner points and their invariant relative relationships, the frame matching problem can be formulated as follows: Given two sets of points $M = {P^M_1, \dots, P^M_m}$ and $N = {P^N_1, \dots, P^N_n}$, where at least $k$ points in set $N$ satisfy the following relationship with set $M$:
\begin{equation}\label{eq:N2M_translation}
    \begin{aligned}
        P^M_{i}=T^M_NP^N_{j}
    \end{aligned}
\end{equation}
where $T^M_N$ represents the relative pose between the coordinate systems of set $M$ and set $N$.

For any point set $M$, upon selecting a point $P^M_i$ from it, we can compute the distance $d^M_{ij}$ and the vector $v = P^M_i - P^M_j$ to other points. Subsequently, the angle between $v$ and $v_x = \begin{matrix} [1 & 0 & 0] \end{matrix}^T$ is calculated and denoted as $a^M_{ij}\in(-\pi,\pi]$. Similar operations can be performed for the other point set $N$. When sets $M$ and $N$ have at least $k$ points that satisfy equation~\eqref{eq:N2M_translation}, the following relationships hold for these $k$ points:
\begin{equation}\label{eq:17}
    \begin{aligned}
        d^M_{ij}&=d^N_{ij}  \\
        diff&=Norm(a^M_{ij}-a^N_{ij}) \\
        Norm(a)&=f(x) = 
        \left\{
                \begin{array}{ll}
                	a \\
                    a - \pi & \mbox{if } a > \pi \\
                    a + \pi & \mbox{if } x < - \pi 
                \end{array}
        \right.
    \end{aligned}
\end{equation}
During the algorithm's actual execution, slight noise and environmental changes may make strict adherence to the length consistency relationship and the angle difference consistency relationship in equation~\eqref{eq:17} challenging. Therefore, we discretize both the distance $d$ and angle $a$ relationships as follows:
\begin{equation}
    \begin{aligned}
        d&=round(\frac{d}{d_{res}})\\
        a&=round(\frac{a}{a_{res}})
    \end{aligned}
\end{equation}
Where $d_{\text{res}}$ represents the allowed maximum distance error, and $a_{\text{res}}$ represents the allowed maximum angle error. For any point $P^M_i$ in the point set $M$, the descriptor $des^M_i = {(d^M_{ij}, a^M_{ij}, j)}$ is obtained using the above method. We arrange all descriptors in the order of $d^M_{ij}$ from smallest to largest into a list $DesM=[des^M_i]$, which represents the set of descriptors of point set $M$. In the matching process, we first find the matches $d^M_{ij}$ and $d^N_{ij}$, and then perform the $\alpha_{ij}$ matching. When the element with the maximum angle difference exceeds the threshold $t$, the matching is considered established, and at least $t$ pairs of matching points are obtained.

In principle, matching all descriptors from set $DesM$ with all descriptors from set $DesN$ can achieve the highest theoretical success rate. However, such an approach leads to a considerably high algorithmic complexity. 

To mitigate this complexity, adopting a randomized strategy proves to be an effective approach. Let the size of point set $M$ be denoted as $m$, and the size of point set $N$ as $n$, with their intersection size being $c$. By randomly selecting $c$ points from point set $M$, the algorithm can successfully accomplish the matching task. If a success rate of no less than $p$ is desired, let the corresponding number of random trials be denoted as $\gamma$. The ensuing relationship can be deduced as follows:
\begin{equation}
    \begin{aligned}
        1-(1-\frac{c}{m})^\gamma>p
    \end{aligned}
\end{equation}
Continued derivation yields the subsequent expression:
\begin{equation}
    \begin{aligned}
        \gamma \geq \lceil \frac{\log{(1-p)}}{\log{(1-\frac{c}{m})}} \rceil.
    \end{aligned}
\end{equation}
In practical operation, successful matching between two frames often results in a relatively substantial intersection $c$. In many instances, it holds true that $\gamma \ll m$, consequently leading to significant savings in terms of $m$-$k$ matchings. To illustrate, when $p=0.95$, $m=20$, and $c=10$, implementing the randomized algorithm facilitates a minimum of 75\% reduction in matching time. This efficiency gain ensures adept matching outcomes while achieving the specified 95\% success rate.

\subsection{Pose Graph Opitimization}
In the domain of global optimization, we employed a pose graph optimization method, akin to the implementation found in VINS\cite{qin_vins-mono_2018}.

By employing the aforementioned matching algorithm, we were able to obtain a minimum of $k$ pairs of matching point pairs denoted as $p^M_{i}, p^N_{j}$. Assuming there are $K$ matching point pairs, solving for the relative pose can be equivalently represented as addressing the ensuing nonlinear least squares problem:
\begin{equation}
    \begin{aligned}
        p^M_n,\theta^M_n=\arg \min{\sum^K_{i=1} 
	\| p^M_{i}-\exp(p^M_N,\theta^M_N)P^N_{j} \|^2
	}.
    \end{aligned}
\end{equation}

\subsection{Probability Grid Map}
The generated map takes the form of a two-dimensional probability grid map. Its approach to updating probabilities closely mirrors that of Cartographer\cite{hess_real-time_2016}, entailing the calculation of hit and miss counts for each individual grid in order to adjust its occupancy probability. Furthermore, the integration of a Gaussian filtering kernel has been introduced to further optimize the grid map.

\section{EXPERIMENT}
\subsection{Experimental Scenarios and Datasets}
This paper utilizes the OpenLORIS-Scene\cite{shi_are_2020} dataset, which comprises visual, IMU, wheel odometry, and 2D LiDAR data, tailored for typical indoor scenarios involving service robots. However, the dataset does not cover 3D LiDAR data, so we mainly compare the algorithm proposed in this paper with mainstream 2D LiDAR SLAM and visual SLAM. The dataset features four distinct indoor scenes: office, home, cafe, and corridor. Notably, the office scene simulates a confined space reminiscent of a small-scale office environment. In the home scene, the robot navigates through a typical household, covering the living room and various rooms. The cafe scene recreates the atmosphere of an operational coffee shop, replete with tables, chairs, and a diverse array of patrons, including pedestrians. Transparent glass elements are also introduced, introducing challenges to the precision of the LiDAR. Lastly, the corridor scene offers a narrow and elongated environment, primarily consisting of a lengthy hallway. However, the presence of transparent windows along the corridor significantly influences the localization accuracy of the 2D LiDAR. Table~\ref{tab:1} provides an overview of trajectory lengths for each scene. It is pertinent to emphasize that all experiments were executed on an environment employing the AMD Ryzen 7 PRO 1700.

\begin{table}[!ht]
    \centering
    \caption{Total trajectory length}
    \begin{tabularx}{\linewidth}{@{\extracolsep{\fill}}ccccc}
    \toprule
        Scenes & Office & Home & Cafe & Corridor \\ \midrule
        Length (m) & 17.96 & 40.06 & 46.84 & 143.20 \\ \bottomrule
    \end{tabularx}
    \label{tab:1}
\end{table}


\captionsetup[subfigure]{font=footnotesize}
\begin{figure}[!ht]
  \centering
  \begin{subfigure}[b]{0.47\linewidth}
    \includegraphics[width=\linewidth]{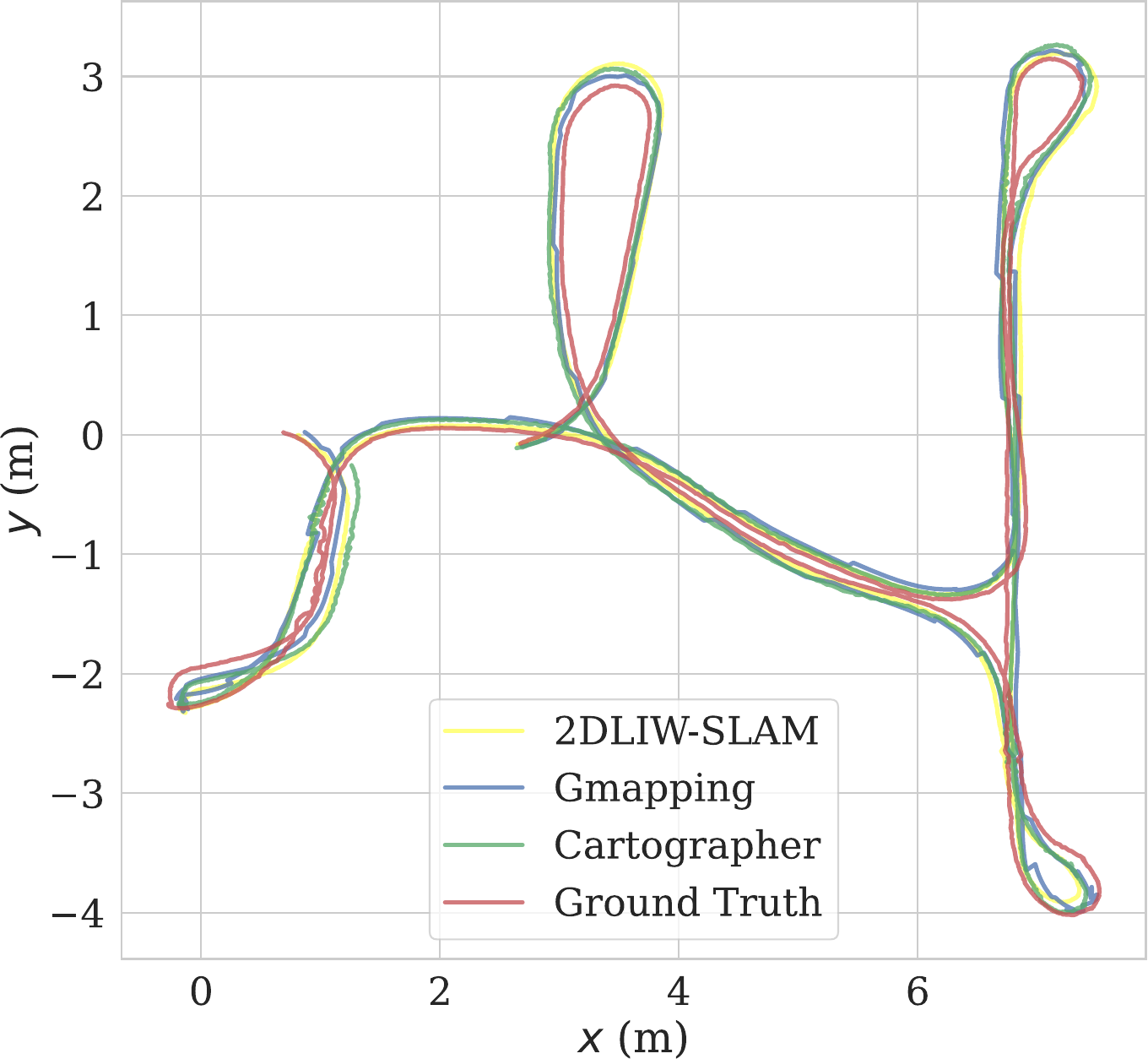}
    \caption{Trajectories in the home scene}
    \label{fig:subimg1}
  \end{subfigure}
  \begin{subfigure}[b]{0.47\linewidth}
    \includegraphics[width=\linewidth,height=0.92\textwidth]{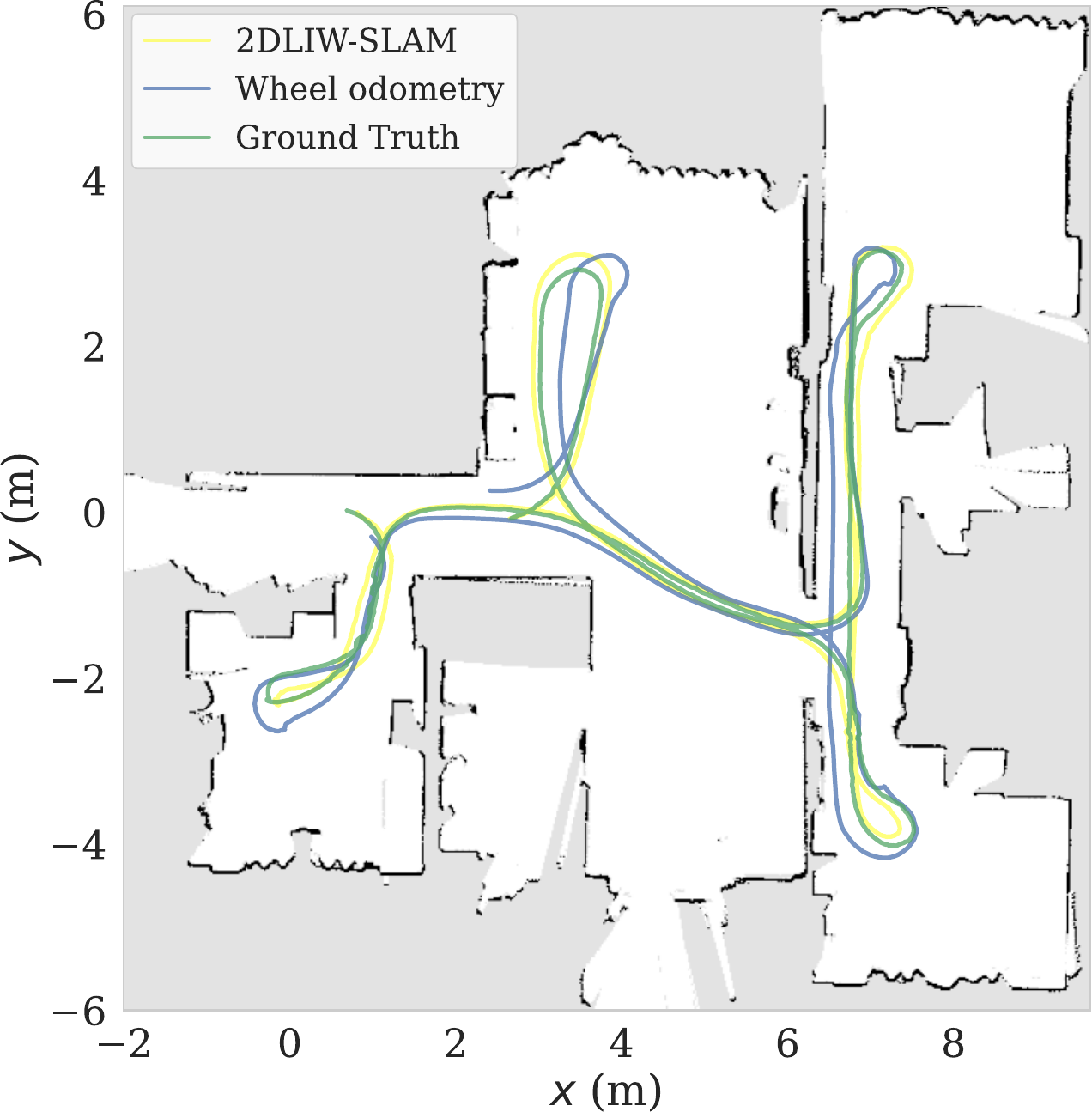}
    \caption{Map constructed by 2D\-LIW}
    \label{fig:subimg2}
  \end{subfigure}
  \caption{The performance of 2DLIW in home scene. 2DLIW can address the drift error in wheel odometry and achieve effective localization and mapping.}
  \label{fig:enter-label}
\end{figure}

\begin{table*}[!ht]
    \centering
    \caption{Absolute pose error compared to previous works.}
    \begin{tabular*}{0.56\linewidth}{@{}ccccc@{\extracolsep{\fill}}}
    \toprule
        Method        & Office         & Home            & Cafe            & Corridor  \\ \midrule
        Gmapping\cite{grisetti_improved_2007}     & 0.063          & \textbf{0.111}  & 0.174           & 0.145   \\ 
        Cartographer\cite{hess_real-time_2016} & 0.096          & 0.114           & 0.176           & 0.129   \\ \midrule
        ORB-SLAM3 (S-I)\cite{camposORBSLAM3AccurateOpenSource2021}    & 0.235          & 0.406           & \textbf{0.164}  & 1.131  \\ 
        VINS-Fusion (S-W)\cite{wuVINSWheels2017}  & 0.323          & 0.570           & 0.401           & 0.504  \\ \midrule
        2DLIW-SLAM w/o loop   & 0.064 & 0.140           & 0.192           & 0.131   \\        
        2DLIW-SLAM   & \textbf{0.060} & 0.115           & 0.170           & \textbf{0.126}   \\ \bottomrule
    \end{tabular*}
    \label{tab:APE}
\end{table*}

\subsection{Pose Error}

To evaluate the accuracy of the odometry, we employed the Root Mean Squared Error (RMSE) of Relative Pose Error (RPE) as the primary assessment metric. For the evaluation of localization precision, we utilized the RMSE of Absolute Pose Error (APE). These aforementioned metrics concurrently consider both translation and rotation errors in the evaluation process.

Table~\ref{tab:RPE} presents the RMSE of RPE, with a pose interval set at 0.1 meters. In the office scene, 2DLIWO exhibits exceptional performance, with an impressively low RMSE of 0.0197, surpassing that of Cartographer and Gmapping and gets the lowest errors across all scenes. Furthermore, Cartographer\cite{hess_real-time_2016} displayed superior tracking performance when contrasted with Gmapping\cite{grisetti_improved_2007}.

\begin{table}[!ht]
    \centering
    \caption{Relative pose error compared to other 2D LiDAR SLAM.}
    \begin{tabularx}{\linewidth}{@{\extracolsep{\fill}}ccccc}
    \toprule
        Method     & Office & Home & Cafe & Corridor  \\ \midrule
        Gmapping\cite{grisetti_improved_2007} & 0.026  & 0.040  & 0.049  & 0.108   \\ 
        Cartographer\cite{hess_real-time_2016} & 0.027  & 0.027  & 0.040  & 0.027   \\ 
        2DLIWO & \textbf{0.020}  & \textbf{0.023}  & \textbf{0.032}  & \textbf{0.026}   \\ \bottomrule
    \end{tabularx}
    \label{tab:RPE}
\end{table}

The RMSE of APE is depicted in table~\ref{tab:APE}. In this experiment, we also tested the open-source visual systems ORB-SLAM3\cite{camposORBSLAM3AccurateOpenSource2021} in Stereo-Inertial (S-I) mode and VINS-Fusion3\cite{wuVINSWheels2017} in Stereo-Wheel (S-W) mode. Among the four scenes, 2DLIW-SLAM performs best in the office and corridor scenes, demonstrating high accuracy in localization. Particularly in the corridor scene, its APE reaches 0.126, significantly outperforming other algorithms. In the home scene, there are a number of walls, so they lack texture features and instead have a rich geometric structure. Therefore, SLAM based on 2D LiDAR achieves higher accuracy, with Gmapping\cite{grisetti_improved_2007} performing the best. Conversely, in the cafe scene, rich texture details enable the visual-IMU based ORB-SLAM3\cite{camposORBSLAM3AccurateOpenSource2021} to achieve better localization accuracy. When faced with challenges such as texture ambiguity or dynamic objects, visual-wheel SLAM performs even worse than 2D LiDAR SLAM. This suggests that 2D LiDAR offers higher robustness in indoor robotics.

\subsection{Loop Detection}
\begin{table}[!ht]
    \centering
    \caption{Global localization in Deutsches Museum Dataset.}
    \begin{tabular*}{\linewidth}{@{\extracolsep{\fill}}cccc}
        \toprule
        \multirow{2}{*}{Methods} & 
        \multirow{2}{*}{\makecell{Time\\(ms)}} & 
        \multirow{2}{*}{\makecell{Translation\\Error (m)}} & 
        \multirow{2}{*}{\makecell{Rotation\\Error (\textdegree)}} \\
         & & & \\ \midrule
        AMCL\cite{foxKLDsamplingAdaptiveParticle2001} & 661.15 & 0.09 & 0.78\\ 
        Cartographer\cite{hess_real-time_2016} & 1021.29 & \textbf{0.06} & \textbf{0.73}\\
        Ours & \textbf{8.99} & 0.07 & 0.74\\ 
         \bottomrule
    \end{tabular*}
    \label{tab:global_localization}
\end{table}

\begin{figure}
    \centering
    \includegraphics[width=0.7\linewidth]{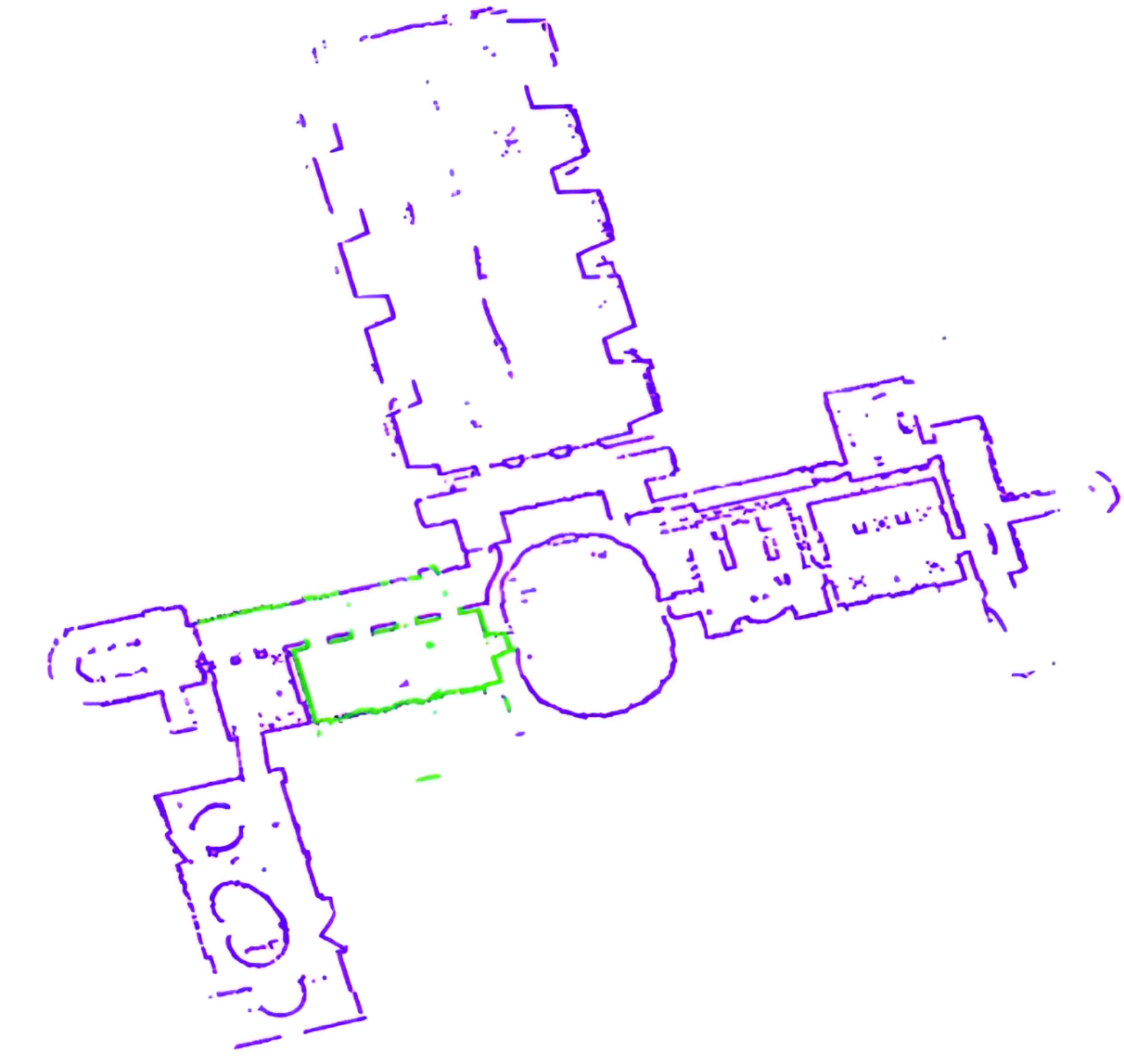}
    \caption{Illustration of successful global localization in Deutsches Museum.}
    \label{fig:match_museum}
\end{figure}

In order to verify the effectiveness of loop detection, we conducted ablation experiments and listed the results in Table~\ref{tab:APE}. The results indicate that after loop detection, optimizing the loop edges helps to improve the SLAM performance.

Furthermore, we isolated the loop matching module and combined it with the Iterative Closest Point (ICP) algorithm to compare global localization effectiveness with AMCL\cite{foxKLDsamplingAdaptiveParticle2001} and Cartographer\cite{hess_real-time_2016} on the Deutsches Museum dataset\cite{hess_real-time_2016}, as depicted in figure~\ref{fig:match_museum}. Four different times and locations were selected as the starting positions of the global localization algorithm. Subsequently, the performance of global localization was evaluated based on the average processing time for each frame and the accuracy of successful localization. Given the vastness of the dataset's scene, to ensure successful localization, we set the number of particles to 20,000 for AMCL, the number of search layers to 7 for Cartographer, the length of the search window edge to 7m, and the search angle to 30°. As shown in table~\ref{tab:global_localization}, our matching optimization algorithm not only achieves high localization accuracy, but also exhibits two orders of magnitude lower time complexity compared to other methods.

Additionally, in table~\ref{tab:loop}, we present the average values of translation error and rotation error for the matched frames in the Openloris dataset. These errors are computed from the difference between the ICP optimization of the matched frames and the positions obtained from front-end tracking.

\subsection{Degeneracy Problem}
\begin{table*}[!ht]
    \centering
    \caption{Relative Pose Error in Degeneracy Problem}
    \begin{tabular*}{\linewidth}{@{\extracolsep{\fill}}cccccccc}
    \toprule
        Method & Max & Mean & Median & Min & Rmse & Sse & Std \\ \midrule
        Gmapping\cite{grisetti_improved_2007} & 0.914 & 0.042 & 0.013 & 0.001 & 0.094 & 7.125 & 0.085 \\ 
        Cartographer\cite{hess_real-time_2016} & 0.511 & 0.018 & 0.009 & 0.001 & 0.042 & 1.673 & 0.038 \\ 
        2DLIW-SLAM & \textbf{0.312} & \textbf{0.012} & \textbf{0.006} & \textbf{0.002} & \textbf{0.027} & \textbf{0.851} & \textbf{0.024} \\ \bottomrule
    \end{tabular*}
    \label{tab:4}
\end{table*}
Within the corridor scene, we imposed a LiDAR measurement range limit of 3 meters. This restriction aimed to simulate an extensive hallway scenario, characterized by similar geometric attributes. Such scenarios often lead to large errors because of the loss of observation information.

\begin{figure}
    \centering
    \includegraphics[width=0.9\linewidth]{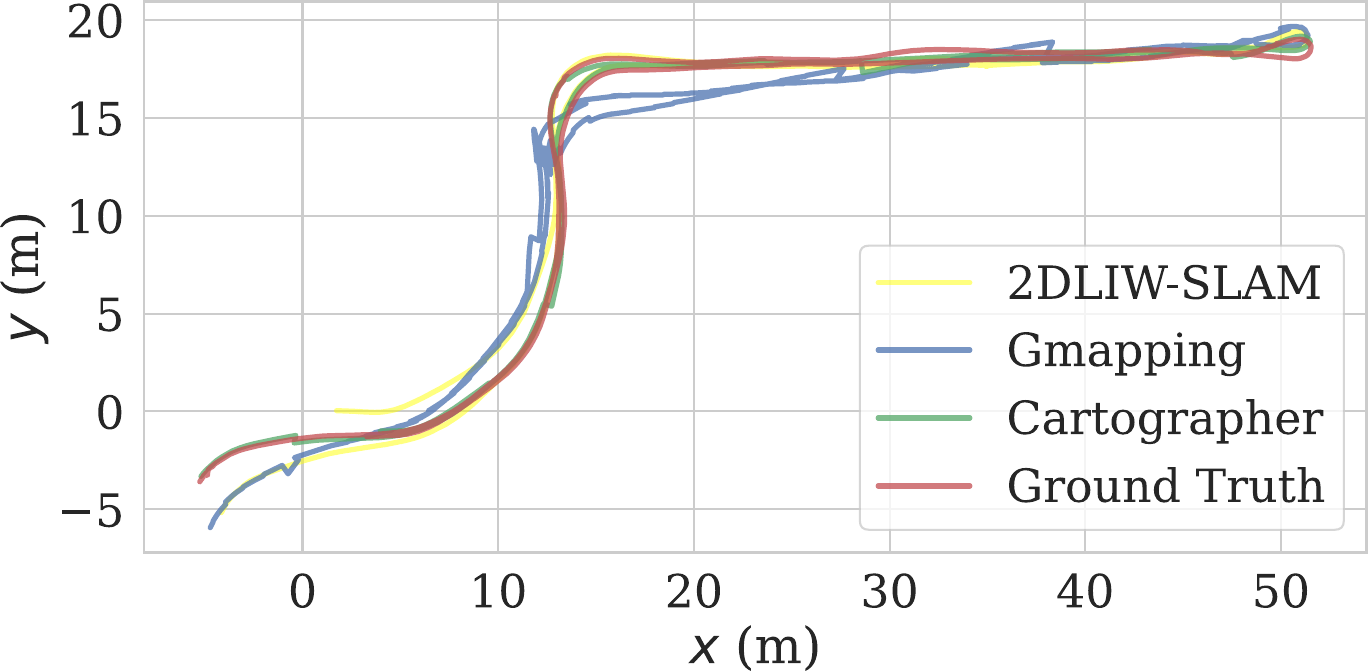}
    \caption{Trajectories in degeneracy problem.}
    \label{fig:degreneracy_trajectory}
\end{figure}

Table~\ref{tab:4} provides an overview of RPE in degeneracy problem, and it's apparent that 2DLIW-SLAM exhibits the most robust tracking performance in this particular scenario and gets the lowest RMSE of 0.027. This can be attributed to the tightly-coupled front-end of 2DLIW-SLAM, which effectively integrates wheel odometry, significantly enhancing the resilience of the front-end odometry.

figure~\ref{fig:degreneracy_trajectory} visually depicts the trajectories of the three algorithms within this challenging scenario. Both 2DLIW-SLAM and Cartographer exhibit effective global localization. However, due to the constraints of the LiDAR's range (limited to 3 meters), loop closure detection through corner feature matching encounters difficulties, resulting in drift errors.

\subsection{Performance of 2DLIW-SLAM}

\begin{figure}
    \centering
    \includegraphics[width=1\linewidth]{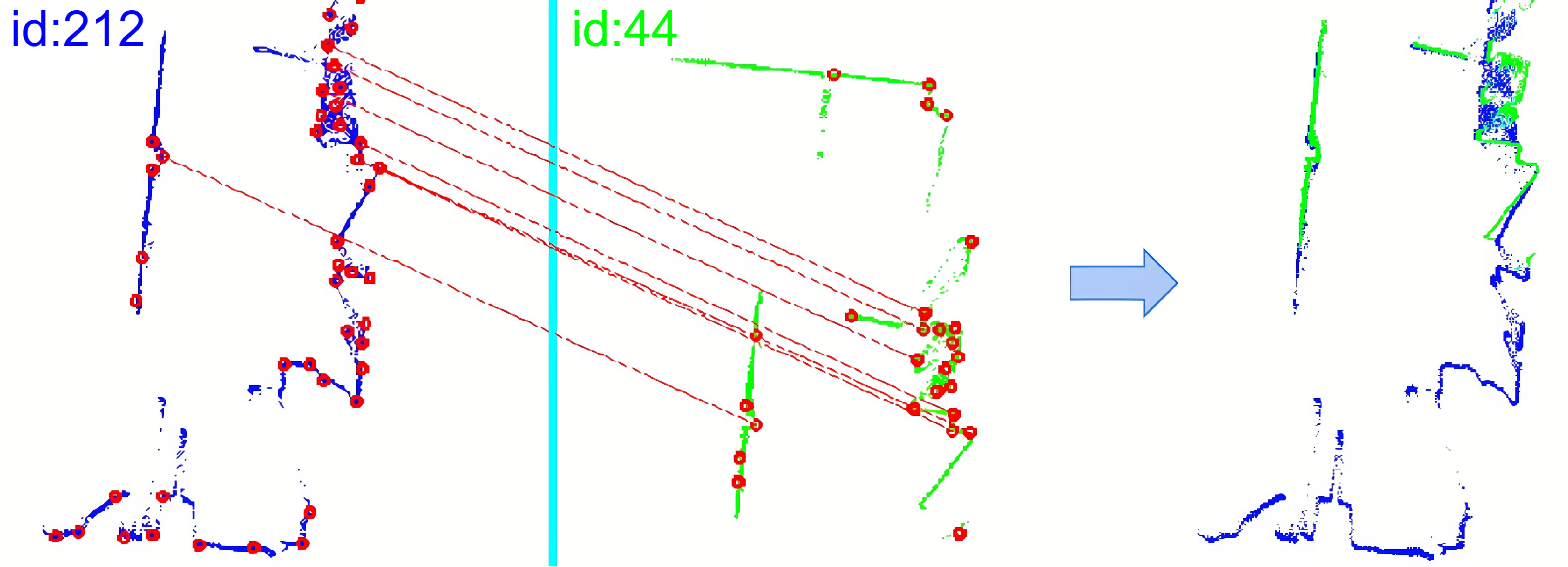}
    \caption{Illustration of successful loop closure matching in the office scene.}
    \label{fig:match}
\end{figure}

\begin{table*}[!ht]
    \centering
    \caption{Performance of Loop Detection}
    \begin{tabular*}{0.9\linewidth}{@{\extracolsep{\fill}}cccccccc}
    \toprule
        \multirow{2}{*}{Scenes} & 
        \multirow{2}{*}{\makecell{Total\\KFs}} & 
        \multirow{2}{*}{\makecell{Corners\\/ KF}} & 
        \multirow{2}{*}{\makecell{Fast Filter\\Ratio}} & 
        \multirow{2}{*}{\makecell{Descriptor\\Match (us)}} & 
        \multirow{2}{*}{\makecell{Frame-to-Frame\\(us)}} & 
        \multirow{2}{*}{\makecell{Translation\\Error (m)}} &
        \multirow{2}{*}{\makecell{Rotation\\Error (\textdegree)}}\\
        & & & & & & & \\ \midrule
        office & 259 & 34.47 & 41.00\% & 1.98 & 394.90 & 0.09 & 0.37 \\ 
        home & 1944 & 18.78 & 93.54\% & 0.19 & 68.70 & 0.03 & 0.44 \\
        cafe & 1342 & 27.96 & 84.09\% & 0.54& 146.74 & 0.08 & 0.75 \\ 
        corridor & 2199 & 18.51 & 99.51\% & 0.03 & 39.65 & 0.42 & 0.30\\ 
         \bottomrule
    \end{tabular*}
    \label{tab:loop}
\end{table*}

\begin{table}[!ht]
    \centering
    \caption{Performance of Tracking}
    \begin{tabular*}{\linewidth}{@{\extracolsep{\fill}}cccc}
        \toprule
        \multirow{2}{*}{Scenes} & 
        \multirow{2}{*}{\makecell{Num. of\\Line Extractions}} & 
        \multirow{2}{*}{\makecell{Num. of\\Line Matches}} & 
        \multirow{2}{*}{\makecell{Time\\(ms)}} \\
         & & & \\ \midrule
        office & 50.09 & 30.84 & 9.47\\ 
        home & 44.26 & 26.36 & 13.42\\
        cafe & 65.46 & 38.32 & 14.66\\ 
        corridor & 42.52 & 34.50 & 12.68\\ 
         \bottomrule
    \end{tabular*}
    \label{tab:tracking}
\end{table}

figure~\ref{fig:subimg1} elegantly portrays the culmination of 2DLIW-SLAM's positioning and mapping prowess in home scene. While wheel odometry substantively contributes to the positioning process, seamless cooperation among various sensors, coupled with adept loop closure detection, effectively counteracts wheel odometry drift errors as figure~\ref{fig:subimg2}, culminating in remarkable positioning and mapping precision.

Table~\ref{tab:tracking} demonstrates that 2DLIW-SLAM meets the real-time requirements for indoor mobile robot applications. By employing the line extraction approach and line-line matching algorithm presented in this paper, a substantial number of lines can be successfully extracted and matched in the majority of cases. When the sliding window size of 5 is used in the tracking process, the average frame rate can exceed 60 frames. Considering that indoor mobile robots typically operate within a frame rate range of 20 to 30 frames, these results satisfactorily meet real-time demands.

A concrete instance of loop closure detection employing this methodology is portrayed in figure~\ref{fig:match}, elegantly illustrating the algorithm's adeptness at identifying meaningful matching relationships between keyframes. Noteworthy is the office scenario, where considerable robot jitter results in substantial environmental shifts. Nevertheless, the algorithm reliably pinpoints loop closures. 

Table~\ref{tab:loop} meticulously documents critical facets of back-end optimization. On one hand, the matching time between individual descriptors is short. Concurrently, frame-to-frame matching durations correspond proportionally to the feature point count. On the other end, the proportion of erroneously filtered matches in the rapid filtering phase inversely correlates with the feature point ratio, a correlation well-aligned with expectations. In practical testing, feature points typically range between 15 and 40 per frame. Consequently, this algorithm convincingly maintains its real-time execution capacity across a majority of scenarios.

\section{Conclusion}
In this paper, we propose a novel, low-cost multi-sensor SLAM system, named 2DLIW-SLAM, designed for indoor mobile robot applications. For the front-end odometry, we incorporate 2D LIDAR, IMU, and wheel odometry, achieving outstanding positioning accuracy through tightly coupled optimization. In the back-end mapping phase, we utilize 2D LiDAR to extract global feature points, implementing efficient loop closure detection. By combining the pose graph, we perform global BA, thereby enhancing the system's positioning accuracy and robustness. Overall, 2DLIW-SLAM meets the requirements for real-time performance and robustness. Also, there are still areas for improvement in the system. For instance, introducing a camera to capture richer visual information and considering the impact of dynamic obstacles on the system. Additionally, the modeling of odometry has not reached an optimal level. In subsequent work, we will make targeted improvements to further enhance the system's performance.
\section*{Data availability statements}
All data that support the findings of this study are included within the article (and any supplementary files).
\section*{Acknowledgments}
This work was supported in part by the National Science Foundation of China under Grant U21A20478, and in part by the Key technology project of Shunde District under Grant 2130218003002.

\section*{Conflict of interest}
The authors declare that they have no conflicts of interest.
\section*{ORCID iDs}
Bin Zhang
~\url{https://orcid.org/0009-0006-2024-7248}
Zexin Peng
~\url{https://orcid.org/0009-0002-2749-2319}
Bi Zeng
~\url{https://orcid.org/0000-0001-8596-8333}
Junjie Lu
~\url{https://orcid.org/0009-0004-2506-7567}

\section*{References}
\bibliographystyle{unsrt}
\bibliography{2DLIW-SLAM}

\begin{thebibliography}{10}

\bibitem{kolhatkar_review_2021}
Chinmay Kolhatkar and Kranti Wagle.
\newblock Review of {SLAM} {Algorithms} for {Indoor} {Mobile} {Robot} with {LIDAR} and {RGB}-{D} {Camera} {Technology}.
\newblock In Margarita~N. Favorskaya, Saad Mekhilef, Rajendra~Kumar Pandey, and Nitin Singh, editors, {\em Innovations in {Electrical} and {Electronic} {Engineering}}, Lecture {Notes} in {Electrical} {Engineering}, pages 397--409, Singapore, 2021. Springer.

\bibitem{hess_real-time_2016}
Wolfgang Hess, Damon Kohler, Holger Rapp, and Daniel Andor.
\newblock Real-time loop closure in {2D} {LIDAR} {SLAM}.
\newblock In {\em 2016 {IEEE} {International} {Conference} on {Robotics} and {Automation} ({ICRA})}, pages 1271--1278, Stockholm, Sweden, May 2016. IEEE.

\bibitem{xu_fast-lio_2021}
Wei Xu and Fu~Zhang.
\newblock {FAST}-{LIO}: {A} {Fast}, {Robust} {LiDAR}-{Inertial} {Odometry} {Package} by {Tightly}-{Coupled} {Iterated} {Kalman} {Filter}.
\newblock {\em IEEE Robotics and Automation Letters}, 6(2):3317--3324, April 2021.

\bibitem{shan_lio-sam_2020}
Tixiao Shan, Brendan Englot, Drew Meyers, Wei Wang, Carlo Ratti, and Daniela Rus.
\newblock {LIO}-{SAM}: {Tightly}-coupled {Lidar} {Inertial} {Odometry} via {Smoothing} and {Mapping}.
\newblock In {\em 2020 {IEEE}/{RSJ} {International} {Conference} on {Intelligent} {Robots} and {Systems} ({IROS})}, pages 5135--5142, October 2020.
\newblock ISSN: 2153-0866.

\bibitem{liuFusionBinocularVision2022}
Zhenbin Liu, Zengke Li, Ao~Liu, Yaowen Sun, and Shiyi Jing.
\newblock Fusion of binocular vision, {{2D}} lidar and {{IMU}} for outdoor localization and indoor planar mapping.
\newblock {\em Measurement Science and Technology}, 34(2):025203, November 2022.

\bibitem{liuVisualSLAMMethod2023}
Fengyu Liu, Yi~Cao, Xianghong Cheng, and Luhui Liu.
\newblock A visual {{SLAM}} method assisted by {{IMU}} and deep learning in indoor dynamic blurred scenes.
\newblock {\em Measurement Science and Technology}, 35(2):025105, November 2023.

\bibitem{heTightlyCoupledLaserinertial2023}
Guojian He, Yisha Liu, and Chengxiang Li.
\newblock Tightly coupled laser-inertial pose estimation and map building based on {{B-spline}} curves.
\newblock {\em Measurement Science and Technology}, 34(12):125130, September 2023.

\bibitem{steux_tinyslam_2010}
Bruno Steux and Oussama~El Hamzaoui.
\newblock {tinySLAM}: {A} {SLAM} algorithm in less than 200 lines {C}-language program.
\newblock In {\em 2010 11th {International} {Conference} on {Control} {Automation} {Robotics} \& {Vision}}, pages 1975--1979, December 2010.

\bibitem{miller_rao-blackwellized_2007}
Isaac Miller and Mark Campbell.
\newblock Rao-{Blackwellized} {Particle} {Filtering} for {Mapping} {Dynamic} {Environments}.
\newblock In {\em Proceedings 2007 {IEEE} {International} {Conference} on {Robotics} and {Automation}}, pages 3862--3869, April 2007.
\newblock ISSN: 1050-4729.

\bibitem{grisetti_improved_2007}
Giorgio Grisetti, Cyrill Stachniss, and Wolfram Burgard.
\newblock Improved {Techniques} for {Grid} {Mapping} {With} {Rao}-{Blackwellized} {Particle} {Filters}.
\newblock {\em IEEE Transactions on Robotics}, 23(1):34--46, February 2007.

\bibitem{konolige_efficient_2010}
Kurt Konolige, Giorgio Grisetti, Rainer Kümmerle, Wolfram Burgard, Benson Limketkai, and Regis Vincent.
\newblock Efficient {Sparse} {Pose} {Adjustment} for {2D} mapping.
\newblock In {\em 2010 {IEEE}/{RSJ} {International} {Conference} on {Intelligent} {Robots} and {Systems}}, pages 22--29, October 2010.
\newblock ISSN: 2153-0866.

\bibitem{durrant-whyte_linear_2012}
Hugh Durrant-Whyte, Nicholas Roy, and Pieter Abbeel.
\newblock A {Linear} {Approximation} for {Graph}-{Based} {Simultaneous} {Localization} and {Mapping}.
\newblock In {\em Robotics: {Science} and {Systems} {VII}}, pages 41--48. MIT Press, 2012.

\bibitem{zhang_loam_2014}
Ji~Zhang and Sanjiv Singh.
\newblock {LOAM}: {Lidar} {Odometry} and {Mapping} in {Real}-time.
\newblock In {\em Robotics: {Science} and {Systems} {X}}. Robotics: Science and Systems Foundation, July 2014.

\bibitem{shan_lego-loam_2018}
Tixiao Shan and Brendan Englot.
\newblock {LeGO}-{LOAM}: {Lightweight} and {Ground}-{Optimized} {Lidar} {Odometry} and {Mapping} on {Variable} {Terrain}.
\newblock In {\em 2018 {IEEE}/{RSJ} {International} {Conference} on {Intelligent} {Robots} and {Systems} ({IROS})}, pages 4758--4765, October 2018.
\newblock ISSN: 2153-0866.

\bibitem{kaess_isam2_2011}
Michael Kaess, Hordur Johannsson, Richard Roberts, Viorela Ila, John Leonard, and Frank Dellaert.
\newblock {iSAM2}: {Incremental} smoothing and mapping with fluid relinearization and incremental variable reordering.
\newblock In {\em 2011 {IEEE} {International} {Conference} on {Robotics} and {Automation}}, pages 3281--3288, May 2011.
\newblock ISSN: 1050-4729.

\bibitem{xuFASTLIO2FastDirect2022}
Wei Xu, Yixi Cai, Dongjiao He, Jiarong Lin, and Fu~Zhang.
\newblock {{FAST-LIO2}}: {{Fast Direct LiDAR-Inertial Odometry}}.
\newblock {\em IEEE Transactions on Robotics}, 38(4):2053--2073, August 2022.

\bibitem{lin_r_2021}
Jiarong Lin, Chunran Zheng, Wei Xu, and Fu~Zhang.
\newblock R {\textasciicircum}2 {LIVE}: {A} {Robust}, {Real}-{Time}, {LiDAR}-{Inertial}-{Visual} {Tightly}-{Coupled} {State} {Estimator} and {Mapping}.
\newblock {\em IEEE Robotics and Automation Letters}, 6(4):7469--7476, October 2021.

\bibitem{lin_r3live_2022}
Jiarong Lin and Fu~Zhang.
\newblock {R3LIVE}: {A} {Robust}, {Real}-time, {RGB}-colored, {LiDAR}-{Inertial}-{Visual} tightly-coupled state {Estimation} and mapping package.
\newblock In {\em 2022 {International} {Conference} on {Robotics} and {Automation} ({ICRA})}, pages 10672--10678, May 2022.

\bibitem{zhong_lvio-sam_2021}
Xinliang Zhong, Yuehua Li, Shiqiang Zhu, Wenxuan Chen, Xiaoqian Li, and Jason Gu.
\newblock {LVIO}-{SAM}: {A} {Multi}-sensor {Fusion} {Odometry} via {Smoothing} and {Mapping}.
\newblock In {\em 2021 {IEEE} {International} {Conference} on {Robotics} and {Biomimetics} ({ROBIO})}, pages 440--445, December 2021.

\bibitem{zhangAccurateRealtimeSLAM2022}
Guangyi Zhang, Tao Zhang, and Chen Zhang.
\newblock Accurate real-time {{SLAM}} based on two-step registration and multimodal loop detection.
\newblock {\em Measurement Science and Technology}, 34(2):025201, November 2022.

\bibitem{liuGLIOTightlyCoupledGNSS2024}
Xikun Liu, Weisong Wen, and Li-Ta Hsu.
\newblock {{GLIO}}: {{Tightly-Coupled GNSS}}/{{LiDAR}}/{{IMU Integration}} for {{Continuous}} and {{Drift-Free State Estimation}} of {{Intelligent Vehicles}} in {{Urban Areas}}.
\newblock {\em IEEE Transactions on Intelligent Vehicles}, 9(1):1412--1422.

\bibitem{songReliableEstimationAutomotive2023}
Rui Song, Yongchun Fang, and Haoqian Huang.
\newblock Reliable {{Estimation}} of {{Automotive States Based}} on {{Optimized Neural Networks}} and {{Moving Horizon Estimator}}.
\newblock {\em IEEE/ASME Transactions on Mechatronics}, 28(6):3238--3249, December 2023.

\bibitem{zhang_LiDAR-imu_2019}
Shaojiang Zhang, Yanning Guo, Qiang Zhu, and Zhiyuan Liu.
\newblock Lidar-{IMU} and {Wheel} {Odometer} {Based} {Autonomous} {Vehicle} {Localization} {System}.
\newblock In {\em 2019 {Chinese} {Control} {And} {Decision} {Conference} ({CCDC})}, pages 4950--4955, June 2019.
\newblock ISSN: 1948-9447.

\bibitem{junior_ekf-loam_2022}
Gilmar P.~Cruz Júnior, Adriano M.~C. Rezende, Victor R.~F. Miranda, Rafael Fernandes, Héctor Azpúrua, Armando~A. Neto, Gustavo Pessin, and Gustavo~M. Freitas.
\newblock {EKF}-{LOAM}: {An} {Adaptive} {Fusion} of {LiDAR} {SLAM} {With} {Wheel} {Odometry} and {Inertial} {Data} for {Confined} {Spaces} {With} {Few} {Geometric} {Features}.
\newblock {\em IEEE Transactions on Automation Science and Engineering}, 19(3):1458--1471, July 2022.

\bibitem{shi_are_2020}
Xuesong Shi, Dongjiang Li, Pengpeng Zhao, Qinbin Tian, Yuxin Tian, Qiwei Long, Chunhao Zhu, Jingwei Song, Fei Qiao, Le~Song, Yangquan Guo, Zhigang Wang, Yimin Zhang, Baoxing Qin, Wei Yang, Fangshi Wang, Rosa H.~M. Chan, and Qi~She.
\newblock Are {We} {Ready} for {Service} {Robots}? {The} {OpenLORIS}-{Scene} {Datasets} for {Lifelong} {SLAM}.
\newblock In {\em 2020 {IEEE} {International} {Conference} on {Robotics} and {Automation} ({ICRA})}, pages 3139--3145, May 2020.
\newblock ISSN: 2577-087X.

\bibitem{kim_scan_2018}
Giseop Kim and Ayoung Kim.
\newblock Scan {Context}: {Egocentric} {Spatial} {Descriptor} for {Place} {Recognition} {Within} {3D} {Point} {Cloud} {Map}.
\newblock In {\em 2018 {IEEE}/{RSJ} {International} {Conference} on {Intelligent} {Robots} and {Systems} ({IROS})}, pages 4802--4809, October 2018.
\newblock ISSN: 2153-0866.

\bibitem{kim_scan_2022}
Giseop Kim, Sunwook Choi, and Ayoung Kim.
\newblock Scan {Context}++: {Structural} {Place} {Recognition} {Robust} to {Rotation} and {Lateral} {Variations} in {Urban} {Environments}.
\newblock {\em IEEE Transactions on Robotics}, 38(3):1856--1874, June 2022.

\bibitem{xiang_fastlcd_2021}
Haodong Xiang, Wenzhong Shi, Wenzheng Fan, Pengxin Chen, Sheng Bao, and Mingyan Nie.
\newblock {FastLCD}: {A} fast and compact loop closure detection approach using {3D} point cloud for indoor mobile mapping.
\newblock {\em International Journal of Applied Earth Observation and Geoinformation}, 102:102430, October 2021.

\bibitem{ma_overlaptransformer_2022}
Junyi Ma, Jun Zhang, Jintao Xu, Rui Ai, Weihao Gu, and Xieyuanli Chen.
\newblock {OverlapTransformer}: {An} {Efficient} and {Yaw}-{Angle}-{Invariant} {Transformer} {Network} for {LiDAR}-{Based} {Place} {Recognition}.
\newblock {\em IEEE Robotics and Automation Letters}, 7(3):6958--6965, July 2022.

\bibitem{cui_bow3d_2023}
Yunge Cui, Xieyuanli Chen, Yinlong Zhang, Jiahua Dong, Qingxiao Wu, and Feng Zhu.
\newblock {BoW3D}: {Bag} of {Words} for {Real}-{Time} {Loop} {Closing} in {3D} {LiDAR} {SLAM}.
\newblock {\em IEEE Robotics and Automation Letters}, 8(5):2828--2835, May 2023.

\bibitem{qin_vins-mono_2018}
Tong Qin, Peiliang Li, and Shaojie Shen.
\newblock {VINS}-{Mono}: {A} {Robust} and {Versatile} {Monocular} {Visual}-{Inertial} {State} {Estimator}.
\newblock {\em IEEE Transactions on Robotics}, 34(4):1004--1020, August 2018.

\bibitem{zeng_orb-slam2_2018}
Fanyu Zeng, Wenchao Zeng, and Yan Gan.
\newblock {ORB}-{SLAM2} with {6DOF} {Motion}.
\newblock In {\em 2018 {IEEE} 3rd {International} {Conference} on {Image}, {Vision} and {Computing} ({ICIVC})}, pages 556--559, June 2018.

\bibitem{loweDistinctiveImageFeatures2004}
David~G. Lowe.
\newblock Distinctive {{Image Features}} from {{Scale-Invariant Keypoints}}.
\newblock {\em International Journal of Computer Vision}, 60(2):91--110, November 2004.

\bibitem{camposORBSLAM3AccurateOpenSource2021}
Carlos Campos, Richard Elvira, Juan J.~G{\'o}mez Rodr{\'i}guez, Jos{\'e}~M. M.~Montiel, and Juan D.~Tard{\'o}s.
\newblock {{ORB-SLAM3}}: {{An Accurate Open-Source Library}} for {{Visual}}, {{Visual}}--{{Inertial}}, and {{Multimap SLAM}}.
\newblock {\em IEEE Transactions on Robotics}, 37(6):1874--1890, December 2021.

\bibitem{wuVINSWheels2017}
Kejian~J. Wu, Chao~X. Guo, Georgios Georgiou, and Stergios~I. Roumeliotis.
\newblock {{VINS}} on wheels.
\newblock {\em 2017 IEEE International Conference on Robotics and Automation (ICRA)}, pages 5155--5162.

\bibitem{foxKLDsamplingAdaptiveParticle2001}
Dieter Fox.
\newblock {{KLD-sampling}}: Adaptive particle filters.
\newblock In {\em Proceedings of the 14th {{International Conference}} on {{Neural Information Processing Systems}}: {{Natural}} and {{Synthetic}}}, {{NIPS}}'01, pages 713--720, Cambridge, MA, USA, January 2001. MIT Press.

\end{thebibliography}

\end{document}